# Path Integral Policy Improvement with Covariance Matrix Adaptation

**Freek Stulp**                                                                                            FREEK.STULP@ENSTA-PARISTECH.FR

Cognitive Robotics, École Nationale Supérieure de Techniques Avancées (ENSTA-ParisTech), Paris
FLOWERS Research Team, INRIA Bordeaux Sud-Ouest, Talence, France

**Olivier Sigaud**                                                                                              OLIVIER.SIGAUD@UPMC.FR

Institut des Systèmes Intelligents et de Robotique, Université Pierre Marie Curie CNRS UMR 7222, Paris

## Abstract

There has been a recent focus in reinforcement learning on addressing continuous state and action problems by optimizing parameterized policies. $PI^2$ is a recent example of this approach. It combines a derivation from first principles of stochastic optimal control with tools from statistical estimation theory. In this paper, we consider $PI^2$ as a member of the wider family of methods which share the concept of probability-weighted averaging to iteratively update parameters to optimize a cost function. We compare $PI^2$ to other members of the same family – Cross-Entropy Methods and CMAES – at the conceptual level and in terms of performance. The comparison suggests the derivation of a novel algorithm which we call $PI^2$-CMA for "Path Integral Policy Improvement with Covariance Matrix Adaptation". $PI^2$-CMA's main advantage is that it determines the magnitude of the exploration noise automatically.

## 1. Introduction

Scaling reinforcement learning (RL) methods to continuous state-action problems, such as humanoid robotics tasks, has been the focus of numerous recent studies (Kober & Peters, 2011; Theodorou et al., 2010). Most of the progress in the domain comes from direct policy search methods based on trajectory rollouts. The recently proposed direct 'Policy Improvement with Path Integrals' algorithm ($PI^2$) is derived from first principles of stochastic optimal control, and is able to outperform gradient-based RL algorithms such as REINFORCE (Williams, 1992) and Natural Actor-Critic (Peters & Schaal, 2008) by an order of magnitude in terms of convergence speed and quality of the final solution (Theodorou et al., 2010).

What sets $PI^2$ apart from other direct policy improvement algorithms is its use of *probability-weighted averaging* to perform a parameter update, rather than using an estimate of the gradient. Interestingly enough, "Covariance Matrix Adaptation – Evolutionary Strategy (CMAES)" and "Cross-Entropy Methods (CEM)" are also based on this concept. It is striking that these algorithms, despite having been derived from very different principles, have converged to almost identical parameter update rules. To the best of our knowledge, this paper is the first to make this relationship between the three algorithms explicit (Section 2). This hinges on 1) re-interpreting CEM as performing probability-weighted averaging; 2) demonstrating that CEM is a special case of CMAES, by setting certain CMAES parameters to extreme values.

A further contribution of this paper is that we conceptually and empirically investigate the differences and similarities between $PI^2$, CEM, and CMAES (Section 3). These comparisons suggest a new algorithm, $PI^2$-CMA, which has the algorithm structure of $PI^2$, but uses covariance matrix adaptation as found in CEM and CMAES. A practical contribution of this paper is that we show how $PI^2$-CMA automatically determines the exploration magnitude, the only parameter which is not straightforward to tune in $PI^2$.

## 2. Background and Related Work

We now describe the CEM, CMAES and $PI^2$ algorithms, and their application to policy improvement.

## 2.1. Cross-Entropy Method (CEM)

Given a $n$-dimensional parameter vector $\boldsymbol{\theta}$ and a cost function $J : \mathbb{R}^n \mapsto \mathbb{R}$, the Cross-Entropy Method (CEM) for optimization searches for the global minimum with the following steps: **Sample** – Take $K$ samples $\boldsymbol{\theta}_{k=1\ldots K}$ from a distribution. **Sort** – Sort the samples in ascending order with respect to the evaluation of the cost function $J(\boldsymbol{\theta}_k)$. **Update** – Recompute the distribution parameters, based only on the first $K_e$ 'elite' samples in the sorted list. **Iterate** – return to the first step with the new distribution, until costs converge, or up to a certain number of iterations.

A commonly used distribution is a multi-variate Gaussian distribution $\mathcal{N}(\boldsymbol{\theta}, \Sigma)$ with parameters $\boldsymbol{\theta}$ (mean) and $\Sigma$ (covariance matrix), such that these three steps are implemented as in (1)-(5). An example of one iteration of CEM is visualized in Figure 1, with a multi-variate Gaussian distribution in a 2D search space[1].

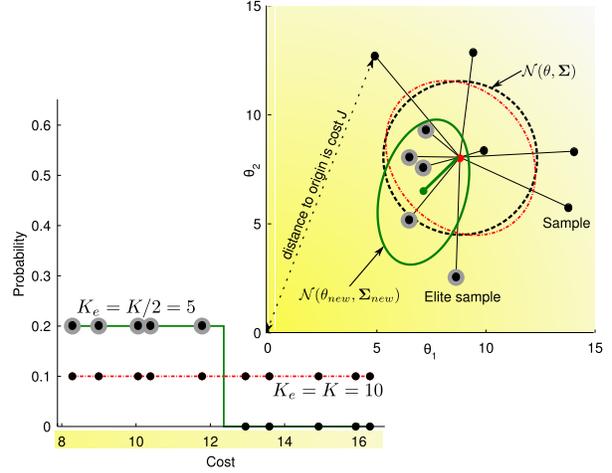

*Figure 1.* Visualization of an update with CEM. The upper right graph shows the 2D parameter space. The cost of a sample is its distance to the origin in Cartesian space. The original multivariate Gaussian distribution $\mathcal{N}([\begin{smallmatrix}8\\8\end{smallmatrix}], [\begin{smallmatrix}9 & 0\\0 & 9\end{smallmatrix}])$ is represented by the dark dashed circle (68% confidence interval). $K = 10$ samples $\boldsymbol{\theta}_k$ are taken from this distribution. The $K_e = 5$ elite samples are used to compute the new Gaussian distribution, which in this case is $\mathcal{N}([\begin{smallmatrix}7.2\\6.4\end{smallmatrix}], [\begin{smallmatrix}3.2 & 1.4\\1.4 & 8.0\end{smallmatrix}])$. The lower left graph shows the mapping from cost to probability, as computed with (12). Note that when $K_e = K$ (red dot-dashed graphs), we are simply estimating the original distribution.

*Cross-Entropy Method (one iteration)*

$$\boldsymbol{\theta}_{k=1\ldots K} \sim \mathcal{N}(\boldsymbol{\theta}, \Sigma) \qquad \textbf{sample} \quad (1)$$
$$J_k = J(\boldsymbol{\theta}_k) \qquad \textbf{eval.} \quad (2)$$
$$\boldsymbol{\theta}_{k=1\ldots K} \leftarrow \texttt{sort } \boldsymbol{\theta}_{k=1\ldots K} \texttt{ w.r.t } J_{k=1\ldots K} \qquad \textbf{sort} \quad (3)$$
$$\boldsymbol{\theta}^{new} = \sum_{k=1}^{K_e} \frac{1}{K_e} \boldsymbol{\theta}_k \qquad \textbf{update} \quad (4)$$
$$\Sigma^{new} = \sum_{k=1}^{K_e} \frac{1}{K_e} (\boldsymbol{\theta}_k - \boldsymbol{\theta})(\boldsymbol{\theta}_k - \boldsymbol{\theta})^{\mathsf{T}} \qquad \textbf{update} \quad (5)$$

Throughout this paper, it is useful to think of CEM as performing *probability-weighted averaging*, where the elite samples have probability $1/K_e$, and the non-elite have probability 0. With these values of $P_k$, (1)-(5) can be rewritten as in the left algorithm in Table 1. Here we use $Q_{K_e/K}$ to denote the $K_e^{\text{th}}$ quantile of the distribution $J_{k=1\ldots K}$. This notation is chosen for brevity; it simply means that in the sorted array of ascending $J_k$, $P_k$ is $1/K_e$ if $K \le K_e$, and 0 otherwise, as in (4). The resulting parameter updates are equivalent to those in (4) and (5), but this representation makes the relation to PI$^2$ more obvious.

**CEM for Policy Improvement.** Because CEM is a very general algorithm, it is used in many different contexts in robot planning and control. CEM for policy optimization was introduced by Mannor et al. (2003). Although their focus is on solving finite small Markov Decision Processes (MDPs), they also propose to use CEM with parameterized policies to solve MDPs with large state spaces. Busoniu et al. (2011) extend this work, and use CEM to learn a mapping from continuous states to discrete actions, where the centers and widths of the basis functions are automatically adapted. The main difference with our work is that we use continuous action spaces of higher dimensionality, and compare CEM to PI$^2$ and CMAES. CEM has also been used in combination with sampling-based motion planning (Kobilarov, 2011). An interesting aspect of this work is that it uses a mixture of Gaussians rather than a single distribution to avoid premature convergence to a local minimum. In (Marin et al., 2011), a CEM is extended to optimize a controller that generates trajectories to any point of the reachable space of the system.

## 2.2. Covariance Matrix Adaptation - Evolution Strategy

The Covariance Matrix Adaptation - Evolution Strategy (Hansen & Ostermeier, 2001) algorithm is very similar to CEM, but uses a more sophisticated method to update the covariance matrix, as listed in Table 2.

---
[1] Note that in (5), the unbiased estimate of the covariance is acquired by multiplying with $\frac{1}{K_e}$, rather than $\frac{1}{K_e-1}$, because we know the true mean to be $\boldsymbol{\theta}$.

| Cross-Entropy Method | Description | $PI^2$ |
|---|---|---|
| | *Exploration Phase* | |
| (6) for $k = 1 \ldots K$ do    $\boldsymbol{\theta}_k \sim \mathcal{N}(\boldsymbol{\theta}, \Sigma)$ | ← loop over trials → <br> ← sample → <br> execute policy → | for $k = 1 \ldots K$ do    $\boldsymbol{\theta}_{k,i=1\ldots N} \sim \mathcal{N}(\boldsymbol{\theta}, \Sigma)$      (7) <br>    $\boldsymbol{\tau}_{k,i=1\ldots N} = \text{executepolicy}(\boldsymbol{\theta}_{k,i=1\ldots N})$      (8) |
| | *Parameter Update Phase* | |
| | loop over time steps → | for $i = 1 \ldots N$ do      (9) |
| for $k = 1 \ldots K$ do | ← loop over trials → |    for $k = 1 \ldots K$ do |
| (10)    $J_k = J(\boldsymbol{\theta}_k)$ | ← evaluate → |      $S_{k,i} \equiv S(\boldsymbol{\tau}_{k,i}) = \sum_{j=i}^{N} J(\boldsymbol{\tau}_{j,k})$      (11) |
| (12)    $P_k = \begin{cases} \frac{1}{K_e} & \text{if } J_k < Q_{K_e/K} \\ 0 & \text{if } J_k > Q_{K_e/K} \end{cases}$ | ← probability → |      $P_{k,i} = \dfrac{e^{-\frac{1}{\lambda} S_{k,i}}}{\sum_{k=1}^{K} [e^{-\frac{1}{\lambda} S_{k,i}}]}$      (13) |
| (14)    $\boldsymbol{\theta}^{new} = \sum_{k=1}^{K} P_k \boldsymbol{\theta}_k$ | ← parameter update → |      $\boldsymbol{\theta}_i^{new} = \sum_{k=1}^{K} P_{k,i} \boldsymbol{\theta}_k$      (15) |
| (16)    $\Sigma^{new} = \sum_{k=1}^{K} P_k (\boldsymbol{\theta}_k - \boldsymbol{\theta})(\boldsymbol{\theta}_k - \boldsymbol{\theta})^\intercal$ | ← covar. matrix adap. → |      $\Sigma_i^{new} = \sum_{k=1}^{K} P_k (\boldsymbol{\theta}_{k,i} - \boldsymbol{\theta})(\boldsymbol{\theta}_{k,i} - \boldsymbol{\theta})^\intercal$      (17) |
| | temporal avg. → | $\boldsymbol{\theta}^{new} = \dfrac{\sum_{i=0}^{N} (N-i) \boldsymbol{\theta}_i^{new}}{\sum_{l=0}^{N} (N-l)}$      (18) |
| | temporal avg. → | $\Sigma^{new} = \dfrac{\sum_{i=0}^{N} (N-i) \Sigma_i^{new}}{\sum_{l=0}^{N} (N-l)}$      (19) |

Table 1. Comparison of the CEM and $PI^2$. This pseudo-code represents one iteration of the algorithm, consisting of an exploration phase and a parameter update phase. Both algorithms iterate these two phases until costs have converged, or up to a certain number of iterations. The green equations – (17) and (19) – are only used in $PI^2$-CMA (to be explained in Section 3.4), and not part of 'standard' $PI^2$.

There are three differences to CEM: • The probabilities in CMAES do not have to be $P_k = 1/K_e$ as for CEM, but can be chosen by the user, as long as the constraints $\sum_{k=1}^{K_e} P_k = 1$ and $P_1 \geq \cdots \geq P_{K_e}$ are met. Here, we use the default suggested by Hansen & Ostermeier (2001), i.e. $P_k = ln(0.5(K+1)) - ln(k)$.
• Sampling is done from a distribution $\mathcal{N}(\boldsymbol{\theta}, \sigma^2 \Sigma)$, i.e. the covariance matrix of the normal distribution is multiplied with a scalar step-size $\sigma$. These components govern the magnitude ($\sigma$) and shape ($\Sigma$) of the exploration, and are updated separately. • For both step-size and covariance matrix an 'evolution path' is maintained ($p_\sigma$ and $p_\Sigma$ respectively), which stores information about previous updates to $\boldsymbol{\theta}$. Using the information in the evolution path leads to significant improvements in terms of convergence speed, because it enables the algorithm to exploit correlations between consecutive steps. For a full explanation of the algorithm we refer to Hansen & Ostermeier (2001).

**Reducing CMAES to CEM.** This is done by setting certain parameters to extreme values: 1) set the time horizon $c_\sigma = 0$. This makes (21) collapse to $\sigma_{new} = \sigma \times \exp(0)$, which means the step-size stays equal over time. Initially setting the step-size $\sigma_{init} = 1$ means $\sigma$ will always be 1, thus having no effect during sampling. 2) For the covariance matrix update, we set $c_1 = 0$ and $c_\mu = 1$. The first two terms of (23) then drop, and what remains is $\sum_{k=1}^{K_e} P_k (\boldsymbol{\theta}_k - \boldsymbol{\theta})(\boldsymbol{\theta}_k - \boldsymbol{\theta})^\intercal$,

*Covariance Matrix Adaptation of CMAES*

$$p_\sigma \leftarrow (1 - c_\sigma) p_\sigma + \sqrt{c_\sigma (2 - c_\sigma) \mu_P} \Sigma^{-1} \frac{\boldsymbol{\theta}^{new} - \boldsymbol{\theta}}{\sigma} \quad (20)$$

$$\sigma_{new} = \sigma \times \exp\left(\frac{c_\sigma}{d_\sigma}\left(\frac{\|p_\sigma\|}{E\|\mathcal{N}(0,I)\|} - 1\right)\right) \quad (21)$$

$$p_\Sigma \leftarrow (1 - c_\Sigma) p_\Sigma + h_\sigma \sqrt{c_\Sigma (2 - c_\Sigma) \mu_P} \frac{\boldsymbol{\theta}^{new} - \boldsymbol{\theta}}{\sigma} \quad (22)$$

$$\Sigma^{new} = (1 - c_1 - c_\mu) \Sigma + c_1 (p_\Sigma p_\Sigma^T + \delta(h_\sigma) \Sigma)$$
$$+ c_\mu \sum_{k=1}^{K_e} P_k (\boldsymbol{\theta}_k - \boldsymbol{\theta})(\boldsymbol{\theta}_k - \boldsymbol{\theta})^\intercal \quad (23)$$

Table 2. The step-size (21) and covariance matrix adaptation (23) update rule of CMAES, which make use of the evolution paths (20) and (22). $\mu_P$ is the variance effective selection mass, with $\mu_P = 1 / \sum_{k=1}^{K_e} P_k^2$. The entire CMAES algorithm is acquired by replacing (16) of CEM in Table 1 with these four equations, and multiplying $\Sigma$ with $\sigma^2$ in (6).

which is equivalent to (16) in CEM, if $P_k$ is chosen as in (12).

**CMAES for Policy Improvement.** Heidrich-Meisner and Igel (2008) use CMAES to directly learn a policy for a double pole-balancing task. Rückstiess et al. (2010) use Natural Evolution Strategies (NES), which has comparable results with CMAES, to di-rectly learn policies for pole balancing, robust standing, and ball catching. The results above are compared with various gradient-based methods, such as REINFORCE (Williams, 1992) and NAC (Peters & Schaal, 2008). To the best of our knowledge, our paper is the first to directly compare CMAES with CEM and PI$^2$. Also, we use Dynamic Movement Primitives as the underlying policy representation, which 1) enables us to scale to higher-dimensional problems, as demonstrated by (Theodorou et al., 2010); 2) requires us to perform temporal averaging, cf. (18) and (19).

### 2.3. Policy Improvement with Path Integrals

A recent trend in reinforcement learning is to use parameterized policies in combination with *probability-weighted averaging*; the PI$^2$ algorithm is a recent example of this approach. Using parameterized policies avoids the curse of dimensionality associated with (discrete) state-action spaces, and using probability-weighted averaging avoids having to estimate a gradient, which can be difficult for noisy and discontinuous cost functions.

PI$^2$ is derived from first principles of optimal control, and gets its name from the application of the Feynman-Kac lemma to transform the Hamilton-Jacobi-Bellman equations into a so-called path integral, which can be approximated with Monte Carlo methods (Theodorou et al., 2010). The PI$^2$ algorithm is listed to the right in Table 1. As in CEM, $K$ samples $\boldsymbol{\theta}_{k=1...K}$ are taken from a Gaussian distribution. In PI$^2$, the vector $\boldsymbol{\theta}$ represents the parameters of a policy, which, when executed, yields a trajectory $\boldsymbol{\tau}_{i=1...N}$ with $N$ time steps. This multi-dimensional trajectory may represent the joint angles of a $n$-DOF arm, or the 3-D position of an end-effector.

So far, PI$^2$ has mainly been applied to policies represented as Dynamic Movement Primitives (DMPs) (Ijspeert et al., 2002), where $\boldsymbol{\theta}$ determines the shape of the movement. Although PI$^2$ searches in the space of $\boldsymbol{\theta}$, the costs are defined in terms of the trajectory $\boldsymbol{\tau}$ generated by the DMP when it is integrated over time. The cost of a trajectory is determined by evaluating $J$ for every time step $i$, where the cost-to-go of a trajectory at time step $i$ is defined as the sum over all future costs $S(\boldsymbol{\tau}_{i,k}) = \sum_{j=i}^{N} J(\boldsymbol{\tau}_{j,k})$, as in (11)[2].

Analogously, the parameter update is applied to every time step $i$ with respect to the cost-to-go $S(\boldsymbol{\tau}_i)$. The probability of a trajectory at $i$ is computed by exponentiating the cost, as in (13). This assigns high probability to low-cost trials, and vice versa. In practice, $-\frac{1}{\lambda}S_{i,k}$ is implemented with optimal baselining as $\frac{-h(S_{i,k}-\min(S_{i,k}))}{\max(S_{i,k})-\min(S_{i,k})}$ cf. (Theodorou et al., 2010).

As can be seen in (15), a different parameter update $\boldsymbol{\theta}_i^{new}$ is computed for each time step $i$. To acquire the single parameter update $\boldsymbol{\theta}^{new}$, the final step is therefore to average over all time steps (18). This average is weighted such that earlier parameter updates in the trajectory contribute more than later updates, i.e. the weight at time step $i$ is $T_i = (N-1)/\sum_{j=1}^{N}(N-1)$. The intuition is that earlier updates affect a larger time horizon and have more influence on the trajectory cost.

PoWeR is another recent policy improvement algorithm that uses probability-weighted averaging (Kober & Peters, 2011). In PoWeR, the immediate costs must behave like an improper probability, i.e. sum to a constant number and always be positive. This can make the design of cost functions difficult in practice; (24) for instance cannot be used with PoWeR. PI$^2$ places no such constraint on the cost function, which may be discontinuous. When a cost function is compatible with both PoWeR and PI$^2$, they perform essentially identical (Theodorou et al., 2010).

## 3. Comparison of PI$^2$, CEM and CMAES

When comparing CEM, CMAES and PI$^2$, there are some interesting similarities and differences. All sample from a Gaussian to explore parameter space – (6) and (7) are identical – and both use probability-weighted averaging to update the parameters – (14) and (15). It is striking that these algorithms, which have been derived within very different frameworks, have converged towards the same principle of probability-weighted averaging.

We would like to emphasize that PI$^2$'s properties follow directly from first principles of stochastic optimal control. For instance, the eliteness mapping follows from the application of the Feymann-Kac lemma to the (linearized) Hamilton Jacobi Bellmann equations, as does the concept of probability-weighted averaging. Whereas in other works the motivation for using CEM/CMAES for policy improvement is based on its empirical performance (Busoniu et al., 2011; Heidrich-Meisner and Igel, 2008; Rückstiess et al., 2010) (e.g. it is shown to outperform a particular gradient-based method), the PI$^2$ derivation (Theodorou et al., 2010) demonstrates that there is a theoretically sound motivation for using methods based on probability-weighted averaging, as this principle follows directly from first principles of stochastic optimal control.

---

[2] For convenience, we abbreviate $S(\boldsymbol{\tau}_{i,k})$ with $S_{i,k}$.

Whereas Section 2 has mainly highlighted the similarities between the algorithms, this section focuses on the differences. Note that any differences between PI$^2$ and CEM/CMAES *in general* also apply to the *specific* application of CEM/CMAES to policy improvement, as done for instance by Busoniu et al. (2011) or Heidrich-Meisner and Igel (2008). Before comparing the algorithms, we first present the evaluation task used in the paper.

### 3.1. Evaluation Task

For evaluation purposes, we use a viapoint task with a 10-DOF arm. The task is visualized and described in Figure 2. This viapoint task is taken from (Theodorou et al., 2010), where it is used to compare PI$^2$ with PoWeR (Kober & Peters, 2011), NAC (Peters & Schaal, 2008), and REINFORCE (Williams, 1992).

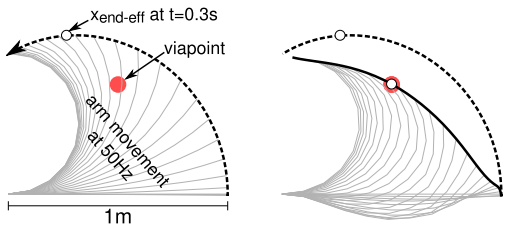

*Figure 2.* The evaluation task. The gray line represents a 10-DOF arm of $1m$ length, consisting of 10 $0.1m$ links. At $t = 0$ the arm is stretched horizontally. Before learning (left figure), each of the joints makes a minimum-jerk movement of 0.5s towards the end position where the end-effector is just 'touching' the $y$-axis. The end-effector path is shown (thick black line), as well as snapshots of the arm posture at 50Hz (thin gray lines). The goal of this task is for the end-effector to pass through the viapoint (0.5,0.5) at $t = 0.3s$, whilst minimizing joint accelerations. The right figure depicts an example of a learned movement.

The goal of this task is expressed with the cost function in (24), where $a$ represents the joint angles, $x$ and $y$ the coordinates of the end-effector, and $D = 10$ the number of DOF. The weighting term $(D+1-d)$ penalizes DOFs closer to the origin, the underlying motivation being that wrist movements are less costly than shoulder movements for humans, cf. (Theodorou et al., 2010).

$$J(\boldsymbol{\tau}_{t_i}) = \delta(t-0.3) \cdot ((\mathbf{x}_t - 0.5)^2 + (y_t - 0.5)^2) + \frac{\sum_{d=1}^{D}(D+1-d)(\ddot{a}_t)^2}{\sum_{d=1}^{D}(D+1-d)} \quad (24)$$

The 10 joint angles trajectories are generated by a 10-dimensional DMP, where each dimension has $B = 5$ basis functions. The parameter vectors $\boldsymbol{\theta}$ (one 1×5 vector for each of the 10 dimensions), are initialized by training the DMP with a minimum-jerk movement. During learning, we run 10 trials per update $K = 10$, where the first of these 10 trials is a noise-free trial used for evaluation purposes. For PI$^2$, the eliteness parameter is $h = 10$, and for CEM and CMAES it is $K_e = K/2 = 5$. The initial exploration noise is set to $\Sigma = 10^4 \mathbf{I}_{B=5}$ for each dimension of the DMP.

### 3.2. Exploration Noise

A first difference between CEM/CMAES and PI$^2$ is the way exploration noise is generated. In CEM and CMAES, time does not play a role, so only one exploration vector $\boldsymbol{\theta}_k$ is generated per trial. In stochastic optimal control, from which PI$^2$ is derived, $\boldsymbol{\theta}_i$ represents a motor command at time $i$, and the stochasticity $\boldsymbol{\theta}_i + \boldsymbol{\epsilon}_i$ is caused by executing command in the environment. When applying PI$^2$ to DMPs, this stochasticity rather represents controlled noise to foster exploration, which the algorithm samples from $\boldsymbol{\theta}_i \sim \mathcal{N}(\boldsymbol{\theta}, \Sigma)$. We call this *time-varying* exploration noise. Since this exploration noise is under our control, we need not vary it at every time step. In the work by Theodorou et al. (2010) for instance, only one exploration vector $\boldsymbol{\theta}_k$ is generated at the beginning of a trial, and exploration is only applied to the DMP basis function that has the highest activation. We call this *per-basis* exploration noise. In the most simple version, called *constant* exploration noise, we sample $\boldsymbol{\theta}_{k,i=0}$ once at the beginning for $i = 0$, and leave it unchanged throughout the execution of the movement, i.e. $\boldsymbol{\theta}_{k,i} = \boldsymbol{\theta}_{k,i=0}$.

The learning curves for these different variants are depicted in Figure 3. We conclude that time-varying exploration convergences substantially slower. Because constant exploration gives the fastest convergence, we use it throughout the rest of the paper.

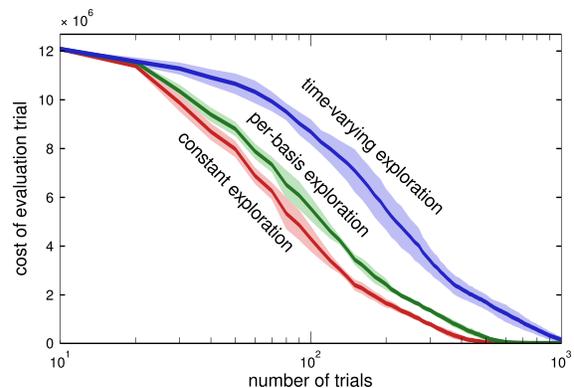

*Figure 3.* Learning curves for time-varying, per-basis and constant exploration.

### 3.3. Definition of Eliteness

In each of the algorithms, the mapping from costs to probabilities is different. CEM implements a cut-off value for 'eliteness': you are either elite ($P_k = 1/K_e$) or not ($P_k = 0$). PI$^2$ rather considers eliteness to be a continuous value that is inversely proportional to the cost of a trajectory. CMAES uses a hybrid eliteness measure where samples have zero probability if they are not elite, and a continuous value which is inverse proportional to the cost if they are elite. The probabilities in CMAES do not have to be $P_k = 1/K_e$ as for CEM, but can be chosen by the user, as long as the constraints $\sum_{k=1}^{K_e} P_k = 1$ and $P_1 \geq \cdots \geq P_{K_e}$ are met. Here, we use the defaults suggested by Hansen & Ostermeier (2001), i.e. $P_k = ln\left(0.5(K+1)\right) - ln(k)$.

These different mappings are visualized in Figure 4. An interesting similarity between the algorithms is that they each have a parameter – $K_e$ in CEM/CMAES, and $h$ in PI$^2$ – that determines how 'elitist' the mapping from cost to probability is. Typical values are $h = 10$ and $K_e = K/2$. These and other values of $h$ and $K_e$ are depicted in Figure 4.

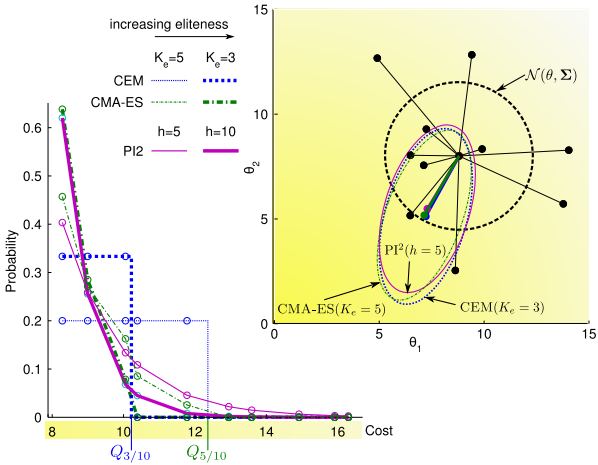

*Figure 4.* Lower left graph: Comparison of the mapping from costs $J_k$ to probabilities $P_k$ for PI$^2$ (with $h = \{10, 5\}$) and CEM/CMAES (with $K_e = \{3, 5\}$). Upper right graph: The updated distributions are very similar with CEM ($K_e = 3$), CMAES ($K_e = 5$) and PI$^2$ ($h = 5$).

The average learning curves in Figure 5 are all very similar except for CEM with $K_e = 5/7$. This verifies the conclusion by Hansen & Ostermeier (2001) that choosing these weights is "relatively uncritical and can be chosen in a wide range without disturbing the adaptation procedure." and choosing the *optimal* weights for a particular problem "only achieves speed-up factors of less than two" when compared with CEM-style weighting where all the weights are $P_k = 1/K_e$. Because choosing the weights is uncritical, we use the PI$^2$ weighting scheme with $h = 10$, the default suggested by Theodorou et al. (2010), throughout the rest of this paper.

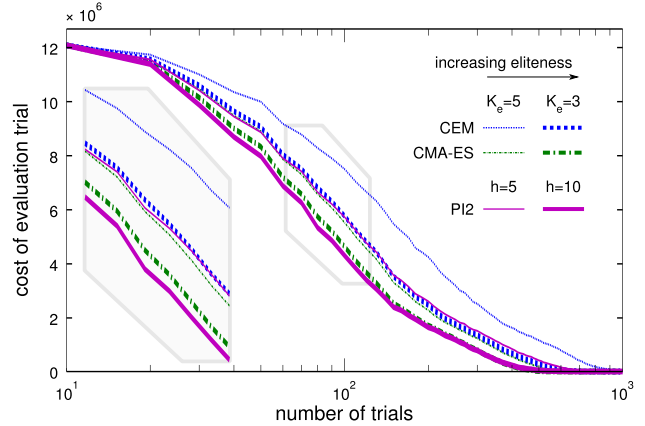

*Figure 5.* Average learning curves for different weighting schemes, averaged over 3 learning sessions. Confidence intervals have been left out for clarity, but are similar in magnitude to those in Figure 3. The inset highlights the similarity with CEM ($K_e = 3$), CMAES ($K_e = 5$) and PI$^2$ ($h = 5$).

### 3.4. Covariance Matrix Adaptation

We now turn to the most interesting and relevant difference between the algorithms. In CEM/CMAES, both the mean and covariance of the distribution are updated, whereas PI$^2$ only updates the mean. This is because in PI$^2$ the shape of the covariance matrix is constrained by the relation $\Sigma = \lambda \mathbf{R}^{-1}$, where $\mathbf{R}$ is the (fixed) command cost matrix, and $\lambda$ is a parameter inversely proportional to the parameter $h$. This constraint is necessary to perform the derivation of PI$^2$ (Theodorou et al., 2010).

In this paper, we choose to ignore the constraint $\Sigma = \lambda \mathbf{R}^{-1}$, and apply covariance matrix updating to PI$^2$. Because a covariance matrix update is computed for each time step $i$ (17), we need to perform temporal averaging for the covariance matrix (19), just as we do for the mean $\boldsymbol{\theta}$. Temporal averaging over covariance matrices is possible, because 1) every positive-semidefinite matrix is a covariance matrix and vice versa 2) a weighted averaging over positive-semidefinite matrices yields a positive-semidefinite matrix (Dattorro, 2011).

Thus, rather than having a fixed covariance matrix, PI$^2$ now adapts $\Sigma$ based on the observed costs for the trials, as depicted in Figure 4. This novel algorithm, which we call PI$^2$-CMA, for "Path Integral

Policy Improvement with Covariance Matrix Adaptation", is listed in Table 1 (excluding the red indices $i = 1 \ldots N$ in (7), and including the green equations (17) and (19)). A second algorithm, PI$^2$-CMAES, is readily acquired by using the more sophisticated covariance matrix updating rule of CMAES. Our next evaluation highlights the main advantage of these algorithms, and compares their performance.

In Figure 6, we compare PI$^2$ (where the covariance matrix is constant[3]) with PI$^2$-CMA (CEM-style covariance matrix updating) and PI$^2$-CMAES (covariance matrix updating with CMAES). Initially, the covariance matrix for each of the 10 DOFs is set to $\Sigma_{init} = \lambda_{init} \mathbf{I}_5$, where 5 is the number of basis functions, and $\lambda_{init} = \{10^2, 10^4, 10^6\}$ determines the initial exploration magnitude. All experiments are run for 200 updates, with $K = 20$ trials per update. We chose a higher $K$ because we are now not only computing an update of the mean of the parameters (a $1 \times 5$ vector for each DOFs), but also its covariance matrix (a $5 \times 5$ matrix), and thus more information is needed per trial to get a robust update (Hansen & Ostermeier, 2001). After each update, a small amount of base level exploration noise is added to the covariance matrix ($\Sigma_{new} \leftarrow \Sigma_{new} + 10^2 \mathbf{I}_5$) to avoid premature convergence, as suggested by Kobilarov (2011).

When the covariance matrices are not updated, the exploration magnitude remains the same during learning, i.e. $\lambda = \lambda_{init}$ (labels Ⓐ in Figure 6), and the convergence behavior is different for the different exploration magnitudes $\lambda_{init} = \{10^2, 10^4, 10^6\}$. For $\lambda_{init} = 10^4$ we have nice convergence behavior Ⓑ, which is not a coincidence – this value has been specifically tuned for this task, and it is the default we have used so far. However, when we set the exploration magnitude very low ($\lambda_{init} = 10^2$) convergence is much slower Ⓒ. When the exploration magnitude is set very high $\lambda_{init} = 10^6$, we get quick convergence Ⓓ. But due to the high stochasticity in sampling, we still have a lot of stochasticity in the cost after convergence in comparison to lower $\lambda_{init}$. This can be seen in the inset, where the $y$-axis has been scaled $\times 20$ for detail Ⓔ.

For PI$^2$-CMA, i.e. *with* covariance matrix updating, we see that the exploration magnitude $\lambda$ changes over time (bottom graph), whereby $\lambda$ is computed as the mean of the eigenvalues of the covariance matrix. For $\lambda_{init} = 10^2$, $\lambda$ rapidly increases Ⓕ until a maximum value is reached, after which it decreases and converges to a value of $10^{2.8}$ Ⓖ. The same holds for $\lambda_{init} = 10^4$, but the initial increase is not so rapid Ⓗ. For $\lambda_{init} = 10^6$, $\lambda$ only decreases Ⓘ, but converges to $10^{2.8}$ as the others.

---

[3]Please note the difference between 1) constant *exploration* as in Section 3.2, where a sampled parameter vector $\boldsymbol{\theta}_k$ is not varied during the movement made in one trial; 2) constant *covariance matrix*, where $\Sigma$ is not updated and thus constant during an entire learning session.

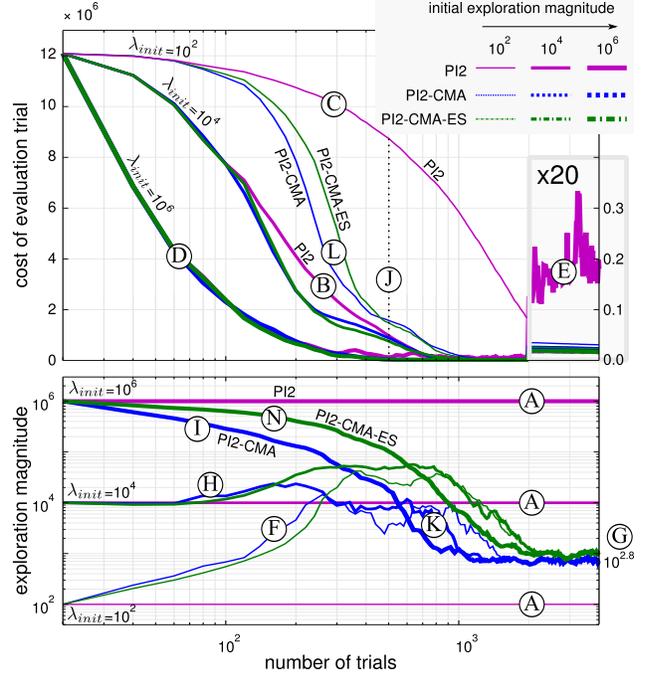

*Figure 6.* Top: Average learning curves with and without covariance matrix updating for different initial exploration magnitudes, averaged over 5 learning sessions. Bottom: The magnitude of the exploration $\lambda$ as learning progresses. Initially $\Sigma_{init} = \lambda_{init} \mathbf{I}_5$ for each DOF.

From these results we derive three conclusions: 1) with PI$^2$-CMA, the convergence speed does not depend as much on the initial exploration magnitude $\lambda_{init}$, i.e. after 500 updates the $\mu \pm \sigma$ cost for PI$^2$-CMA over all $\lambda_{init}$ is $10^5 \cdot (8 \pm 7)$, whereas for PI$^2$ without covariance matrix updating it is $10^5 \cdot (35 \pm 43)$ Ⓙ. 2) PI$^2$-CMA automatically increases $\lambda$ if more exploration leads to quicker convergence ⒻⒽ. 3) PI$^2$-CMA automatically decreases $\lambda$ once the task has been learned ⒼⓀ. Note that 2) and 3) are emergent properties of covariance matrix updating, and has not been explicitly encoded in the algorithm. In summary, PI$^2$-CMA is able to find a good exploration/exploitation trade-off, independent of the initial exploration magnitude.

This is an important property, because setting the exploration magnitude by hand is not straightforward, because it is highly task-dependent, and might require several evaluations to tune. One of the main contributions of this paper is that we demonstrate how using *probability-weighted averaging to update the covariance matrix* (as is done in CEM) allows PI$^2$ to *autonomously tune the exploration magnitude* – the user thus no longer needs to tune this parameter. The only remaining parameters of PI$^2$ are $K$ (number of trials per update) and $h$ (eliteness parameter), but choosing them is not critical. Although an initial $\Sigma$ must be given, Figure 6 shows that with an initial exploration magnitude two orders of magnitude higher/lower than a tuned value, PI$^2$-CMA still converges to the same cost and exploration magnitude, with only slight differences in the initial speed of convergence.

When comparing PI$^2$-CMA and PI$^2$-CMAES, we only see a very small difference in terms of convergence when the initial exploration is low $\lambda_{init} = 10^2$ ⓛ. This is because the covariance update rule of CMAES is damped, (21) and (23), and it makes more conservative updates than CEM, cf. ⓘ and ⓝ. In our experiments, PI$^2$-CMAES uses the default parameters suggested by Hansen & Ostermeier (2001). We have tried different parameters for PI$^2$-CMAES, the conclusion being that the best parameters are those that reduce CMAES to CEM, cf. Section 2.2. In general, we do not claim that PI$^2$-CMAES outperforms PI$^2$, and Hansen & Ostermeier (2001) also conclude that there are tasks where CMAES has identical performance to simpler algorithms. Our results on comparing PI$^2$-CMAES and PI$^2$-CMA are therefore not conclusive. An interesting question is whether typical cost functions found in robotics problems have properties that do not allow CMAES to leverage the advantages it has on benchmark problems used in optimization.

## 4. Conclusion

In this paper, we have scrutinized the recent state-of-the-art direct policy improvement algorithm PI$^2$ from the specific perspective of belonging to a family of methods based on the concept of probability-weighted averaging. We have discussed similarities and differences between three algorithms in this family, being PI$^2$, CMAES and CEM. In particular, we have demonstrated that using probability-weighted averaging to update the covariance matrix, as is done in CEM and CMAES, allows PI$^2$ to autonomously tune the exploration magnitude. The resulting algorithm PI$^2$-CMA shows more consistent convergence under varying initial conditions, and alleviates the user from having to tune the exploration magnitude parameter by hand. We are currently applying PI$^2$-CMA to challenging tasks on a physical humanoid robot. Given the ability of PI$^2$ to learn complex, high-dimensional tasks on real robots (Stulp et al., 2011), we are confident that PI$^2$-CMA can also successfully be applied to such tasks.

## Acknowledgments

We thank the reviewers for their constructive suggestions for improvement of the paper. This work is supported by the French ANR program (ANR 2010 BLAN 0216 01), more at http://macsi.isir.upmc.fr